\documentclass[10pt,twocolumn,letterpaper]{article}

\usepackage{iccv}
\usepackage{times}
\usepackage{epsfig}
\usepackage{graphicx}
\usepackage{amsmath}
\usepackage{amssymb}
\usepackage[table]{xcolor}
\usepackage{graphicx}
\usepackage{booktabs}
\usepackage[T1]{fontenc}
\usepackage[utf8]{inputenc}
\usepackage[english]{babel}
\usepackage{multirow}
\usepackage{multicol}
\usepackage{caption}
\usepackage{subcaption}
\usepackage{amsmath}
\usepackage{amssymb}
\usepackage{capt-of}
\usepackage[export]{adjustbox}
\usepackage{pifont}

\usepackage{makecell}

\usepackage[pagebackref=true,breaklinks=true,letterpaper=true,colorlinks,bookmarks=false]{hyperref}

\iccvfinalcopy 

\def\iccvPaperID{****} 

\ificcvfinal\fi

\begin{document}

\title{Self-Supervised 3D Monocular Object Detection by Recycling Bounding Boxes}
\author{Sugirtha T$^{1}$, Sridevi M$^{1}$, Khailash Santhakumar$^{2}$, Hao Liu$^{3}$\\%
B Ravi Kiran$^{3}$,  Thomas Gauthier$^3$ and Senthil Yogamani$^{4}$ \\
$^{1}$NIT Tiruchirappalli, India \quad
$^{2}$SASTRA University, India \quad
$^{3}$Navya, France \quad
$^{4}$Valeo, Ireland
}
\maketitle
\ificcvfinal\thispagestyle{empty}\fi

\begin{abstract}
   Modern object detection architectures are moving towards employing self-supervised learning (SSL) to improve performance detection with related pretext tasks. Pretext tasks for monocular 3D object detection have not yet been explored yet in literature. The paper studies the application of established self-supervised bounding box recycling by labeling random windows as the pretext task. The classifier head of the 3D detector is trained to classify random windows containing different proportions of the ground truth objects, thus handling the foreground-background imbalance. We evaluate the pretext task using the RTM3D detection model as baseline, with and without the application of data augmentation. We demonstrate  improvements of between  2-3 \% in mAP 3D and 0.9-1.5 \% BEV scores using SSL over the baseline scores. We propose the inverse class frequency re-weighted (ICFW) mAP score that highlights improvements in detection for low frequency classes in a class imbalanced dataset with long tails. We demonstrate improvements in ICFW both mAP 3D and BEV scores to take into account the class imbalance in the KITTI validation dataset. We see 4-5 \% increase in ICFW metric with the pretext task.
\end{abstract}

\section{Introduction}

3D object detection is a crucial perception task in modern autonomous driving applications, used upstream for scene understanding, object tracking and trajectory prediction and decision making. Initially, autonomous cars are equipped with LiDAR sensors and most 3D detectors rely on LiDAR data to perform 3D object detection. LiDAR provides precise distance measurement which makes it feasible to detect accurate 3D bounding boxes. But, they are expensive to be deployed in autonomous cars. Recent autonomous cars use single monocular camera and hence monocular 3D object detection (3D OD) became a research focus in computer vision community. 

\begin{figure*}[!tbh]
\begin{center}
\includegraphics[width=0.85\textwidth]{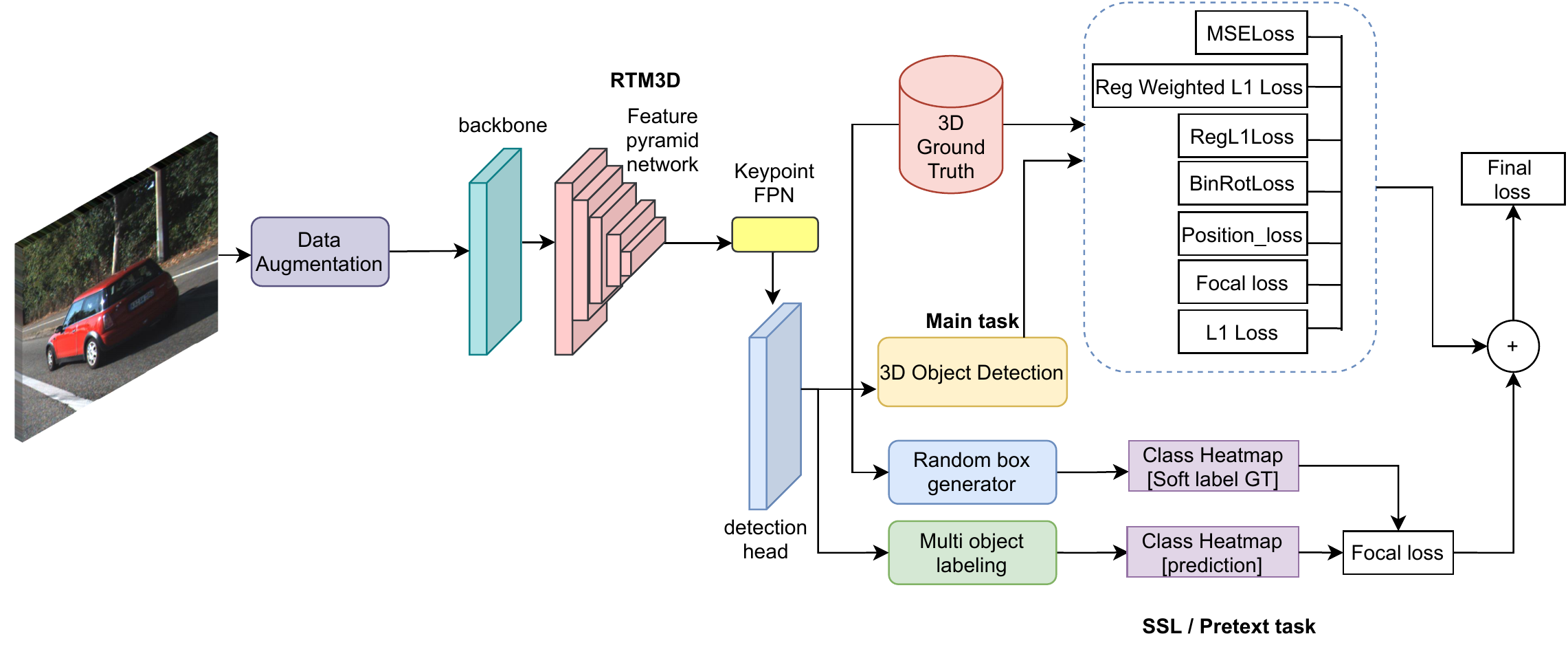}
\caption{Self supervised learning setup with monocular 3D object detection as main task, and multi-object labeling as pretext task. Following authors in \cite{lee2019multi}, we train the main and pretext task in a multi-task learning setup by summing the two losses with equal static weights.}
\label{fig:ssl}
\end{center}
\end{figure*}

RTM3D \cite{rtm3deccv2020} is a monocular 3D object detector based on the CenterNet architecture \cite{Zhou2019ObjectsAP}, we shall use this model as our baseline model to evaluate SSL methods for object detection. In this paper we evaluate the use of multi-object labeling pretext task proposed by authors in \cite{lee2019multi} as self-supervision to improve the 3D monocular object detection.

3D monocular detection requires an expensive annotation process. Self supervised learning methods provide auxiliary or pretext tasks that use cheaply available labels, to help the downstream primary task of monocular 3D-OD. In summary, the contributions of our paper are as follows: 1. We evaluate the performance of baseline RTM3D detector with self-supervised multi-object labeling pretext task under 2 settings (i) different number of random windows as a hyper-parameter (ii) Under the use of data augmentations along with self-supervision. 2. Propose the class frequency sensitive detection score (ICFW) that measures improvement in low frequency critical classes i.e. in pedestrian and cyclist classes.

Extensive analysis on KITTI dataset \cite{kittidataset2012} demonstrates that our proposed data augmentations improve performance under various conditions: occlusions, contrasted/shadowed pixels, changing the diversity of viewpoints of objects seen in the dataset.

\section{Related Work}

2D OD on image plane is inadequate for reliable autonomous driving scenario because it does not provide an accurate estimation of 3D objects sizes and space localization. In other words, 2D OD methods have limited performance in following scenarios namely  occlusion,  object pose estimation and 3D position information. A 3D bounding box provides precise information about the size of the object and its position in 3D space.

\noindent
\textbf{3D Object Detection :}
 3D OD methods are usually part of the 4 following categories: (i) 2D proposal generation \cite{Law_2018_ECCV}, (ii) Geometric constraints \cite{Zhou_2019_CVPR}, (iii) Key-points detection \cite{Zhou2019ObjectsAP} or (iv) Direct 3D proposal generation \cite{Li2020Monocular3D}.  RTM3D\cite{rtm3deccv2020} and its extension KM3D \cite{Li2020Monocular3D} using CenterNet \cite{Zhou2019ObjectsAP} to regress a set of 9 projected keypoints corresponding to a 3D cuboid in image space (8 vertices of the cuboid and its center). They also perform direct regression for the object's distance, size and orientation. These values are then used for offline initialization of an optimizer to estimate 3D bounding boxes under geometric constraints.
 CenterNet provides basic data augmentation such as affine 
 transformations (shifting, scaling) and random horizontal flipping. Over
 these, KM3D \cite{Li2020Monocular3D} adds coordinate independent augmentation via random color
 jittering. 

SMOKE \cite{liu2020smoke} regresses 3D bounding box directly from image plane which eliminates 2D bounding box regression. It represents an object by a single keypoint and these keypoints are projected as 3D center of each object. Mono3D \cite{Chen_2016_CVPR} is a region proposal based method that uses semantics, object contours and location priors to generate 3D anchors. It generates proposal by performing exhaustive search on 3D space and uses non-maximal suppression for filtering. SMOKE augments the training samples with horizontal flipping, scaling and shifting. GS3D \cite{Li_2019_CVPR} predicts the guidance of cuboid and performs feature extraction by projecting region of guidance. GS3D performs monocular 3D detection without augmenting the training data.

\subsection{Self-supervised learning} 
Supervised learning requires human endeavour to create high quality annotations whereas self-supervised learning (SSL) creates labels by their own models without need for human labour. In the area of computer vision, various clues like optical flow \cite{Pathak2017LearningFB}, tracking \cite{Singh2016TrackAT}, inpainting \cite{Pathak2016ContextEF}, sound \cite{Owens2016AmbientSP} and colorization \cite{Zhang2016ColorfulIC} are being utilized as a pretext task which help the primary task generalize better. 
Authors \cite{Noroozi2016UnsupervisedLO} train the network to solve jigsaw puzzles and fine tune it for object localization and detection where \cite{Jenni2018SelfSupervisedFL} differentiates real and artifact images and transfer it to object detection.
Authors in \cite{Zhang2016ColorfulIC} trained the model for colorization purpose and modulated it for object detection. Authors in \cite{Beker2020MonocularDR} 
estimate 3D object properties such as location, dimension etc. via SSL and differentiable rendering, which eliminates the need for 3D annotations.
\cite{lee2019multi} created three auxiliary tasks which reused bounding box labels for self supervision in order to improve 2D object detection performance.  
In this paper, we reiterate this recycling bounding box task \cite{lee2019multi}, reusing the same method to achieve better 3d-monocular object detection.

\section{SSL for Object detection}

Authors in \cite{lee2019multi}, discuss the use of three different pretext tasks to improve the performance of the main/downstream object detection task. Key pretext tasks include : 

\begin{figure}
    \centering
    \includegraphics[width=\linewidth]{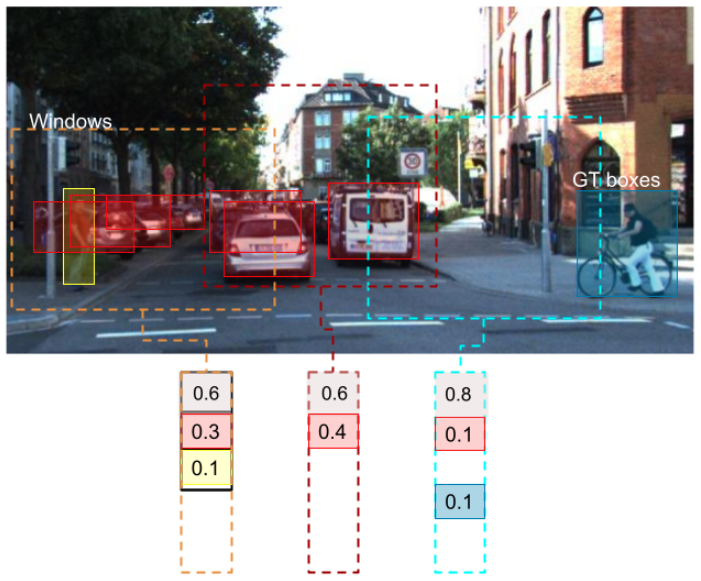}
    \caption{Visualization for 3 different random windows selected from the input image domain, along with their soft label generated by using the proportion of different classes \& background within each window.}
    \label{fig:mol_ssl}
\end{figure}

\begin{itemize}
    \item \textbf{Multi-Object Labeling (MOL)} : To handle the imbalance in foreground and background w.r.t an object detection task, authors propose the usage of a random window ($W$) with partial or complete intersection with objects in the given sample (image $I$, bounding boxes $\{B\}$) pair to artificially increase the number of bounding boxes. The multi-object soft label corresponding to the random window assigns area ratios of each class’s GT boxes $B$ within the random window $W$. Only the classifier head of Faster R-CNN and R-FCN models were trained in a multi-task setting (detection as the main task, random window classification as an auxiliary task). We have demonstrated this on a KITTI image sample in figure \ref{fig:mol_ssl}.
    \item \textbf{Closeness \& FG segmentation} : Authors in  \cite{lee2019multi} also proposed a closeness label that measures the distances from the center of a GT box to those of other GT boxes. While the segmentation task performs binary foreground/background segmentation, where foreground is created from union of regions of all bounding boxes.
\end{itemize}
In our study, we only evaluate the performance of the MOL pretext task.

\subsection{Data augmentations}
\begin{figure*}
    \centering
    \includegraphics[width=0.9\linewidth]{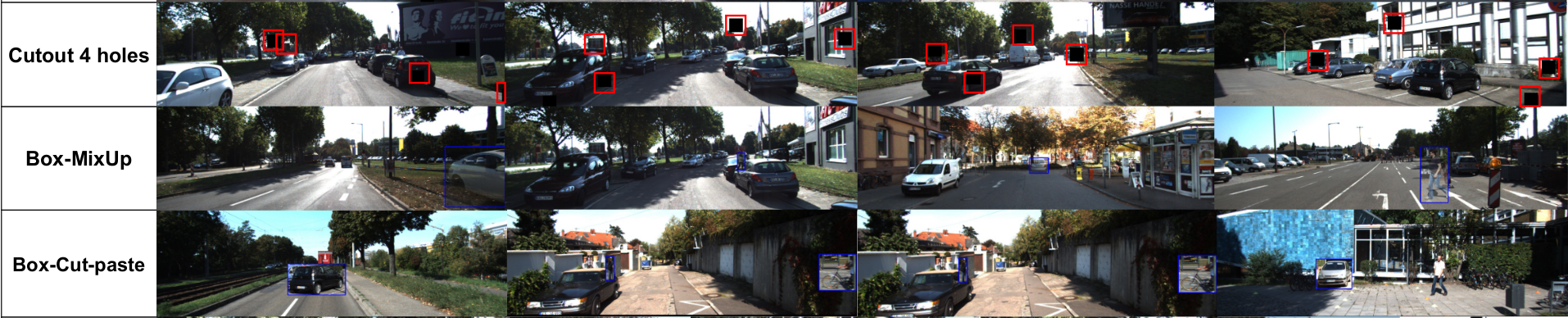}
    \caption{The three data augmentations : Box-mixup, Box-Cutpaste and Cutout that were used along with MOL pretext tasks.}
    \label{fig:data_aug_3}
\end{figure*}

\textbf{Box-Mixup :}
Motivated by the work on Mixup data augmentation \cite{Zhang2018mixupBE}  Box-MixUp is proposed to augment an image with object patches from other images, thus providing the same advantages of MixUp but localized over multiple regions in the image. The augmented sample can be expressed as:
\begin{align}
\begin{split}
\widetilde{x} & = (0.5 x_A + 0.5 x_B) \cdot M_B + x_A \cdot (1-M_B)\\
\widetilde{y} & = y_A \cup y_B \label{eq:1}
\end{split}
 \end{align}

\textbf{Box-Cut-Paste :}
Following cut-paste data augmentation \cite{Dwibedi2017CutPA} in Box Cut-Paste, we paste the pixels under the bounding box mask from one image onto the reference image. This can be expressed as :
\begin{align}
\begin{split}
\widetilde{x} & = x_A \cdot (1-M_B) + x_B \cdot M_B \\
\widetilde{y} & = y_A \cup y_B \label{eq:2}
\end{split}
\end{align}

The augmentations are demonstrated in figure \ref{fig:data_aug_3}.
\section{Experiments}
\textbf{Baseline Model} :   RTM3D \cite{rtm3deccv2020} is a real-time network that is based on the CenterNet architecture, which enables both fast training cycles and small inference time. Authors have already provided a  set of baseline data augmentation which include flip, affine transformations, stereo dataset augmentation using the left/right images in the KITTI dataset.

In our study we present the following comparisons. The RTM3D baseline model trained with multi-object labeling as pretext task in two settings: 1. with pretext task alone 2. pretext task along with data augmentation.

\textbf{SSL task: Multi-Object-Labelling}:
Self supervised learning setup for our evaluation is shown in Fig. \ref{fig:ssl}.
We evaluate the performance of the Multi-Object-Labelling pretext task proposed by authors in \cite{lee2019multi}. We evaluated the performance at different number of random windows as a hyper-parameter :  two, four, eight, and six-teen random windows. We evaluated uniform different distribution of scales of random windows.


\section{Results}

\begin{table*}[hbt!]
\caption{mAP and ICFW mAP scores for both 3D and BEV detection bounding boxes. Green refers to positive gains, while red refers to negative drops in performance over the baseline.}
\label{tab:mAP}
\resizebox{0.95\textwidth}{!}{\begin{minipage}{\textwidth}
\begin{tabular}{|l|c|c|c|c|l|c|l|c|l|}
\hline
\textbf{IoU=0.5}                    & \textbf{mAP$_{2D}$}                & \textbf{mAP$_{BEV}$}               & \textbf{mAP$_{3D}$}                & \multicolumn{2}{c|}{\textbf{ICFW mAP$_{2D}$}}           & \multicolumn{2}{c|}{\textbf{ICFW mAP$_{BEV}$}}          & \multicolumn{2}{c|}{\textbf{ICFW mAP$_{3D}$}}           \\ \hline
\textbf{Baseline (B)}               & \cellcolor[HTML]{D9D9D9}41.44 & \cellcolor[HTML]{D9D9D9}21.17 & \cellcolor[HTML]{D9D9D9}19.12 & \multicolumn{2}{c|}{\cellcolor[HTML]{D9D9D9}33}    & \multicolumn{2}{c|}{\cellcolor[HTML]{D9D9D9}15.1}  & \multicolumn{2}{c|}{\cellcolor[HTML]{D9D9D9}14.65} \\ \hline
\multicolumn{10}{|c|}{\textbf{Self-Supervised Learning (SSL)}}                                                                                                                                                                                                                                                                \\ \hline
\textbf{B + 8W}                     & \cellcolor[HTML]{B7E1CD}0.85  & \cellcolor[HTML]{B7E1CD}0.53  & \cellcolor[HTML]{B7E1CD}0.46  & \multicolumn{2}{c|}{\cellcolor[HTML]{B7E1CD}0.83}  & \multicolumn{2}{c|}{\cellcolor[HTML]{B7E1CD}0.7}   & \multicolumn{2}{c|}{\cellcolor[HTML]{B7E1CD}0.54}  \\ \hline
\textbf{B + 16W}                    & \cellcolor[HTML]{B7E1CD}0.59  & \cellcolor[HTML]{F4C7C3}-0.75 & \cellcolor[HTML]{F4C7C3}-0.59 & \multicolumn{2}{c|}{\cellcolor[HTML]{B7E1CD}0.57}  & \multicolumn{2}{c|}{\cellcolor[HTML]{F4C7C3}-1.88} & \multicolumn{2}{c|}{\cellcolor[HTML]{F4C7C3}-1.73} \\ \hline
\textbf{B + 32W}                    & \cellcolor[HTML]{B7E1CD}1.4   & \cellcolor[HTML]{B7E1CD}0.29  & \cellcolor[HTML]{B7E1CD}0.12  & \multicolumn{2}{c|}{\cellcolor[HTML]{B7E1CD}1.75}  & \multicolumn{2}{c|}{\cellcolor[HTML]{B7E1CD}0.12}  & \multicolumn{2}{c|}{\cellcolor[HTML]{F4C7C3}-0.17} \\ \hline
\multicolumn{10}{|c|}{\textbf{Data Augmentation (DA)}}                                                                                                                                                                                                                                             \\ \hline
\textbf{B + Cutout4}                & \cellcolor[HTML]{F4C7C3}-0.91 & \cellcolor[HTML]{B7E1CD}0.11  & \cellcolor[HTML]{F4C7C3}-0.71 & \multicolumn{2}{c|}{\cellcolor[HTML]{F4C7C3}-2.79} & \multicolumn{2}{c|}{\cellcolor[HTML]{B7E1CD}0.15}  & \multicolumn{2}{c|}{\cellcolor[HTML]{F4C7C3}-0.54} \\ \hline
\textbf{B + BoxMixup}               & \cellcolor[HTML]{B7E1CD}0.39  & \cellcolor[HTML]{B7E1CD}0.29  & \cellcolor[HTML]{B7E1CD}0.21  & \multicolumn{2}{c|}{\cellcolor[HTML]{B7E1CD}0.53}  & \multicolumn{2}{c|}{\cellcolor[HTML]{B7E1CD}0.12}  & \multicolumn{2}{c|}{\cellcolor[HTML]{B7E1CD}0.04}  \\ \hline
\textbf{B + Cutpaste}               & \cellcolor[HTML]{B7E1CD}1.63  & \cellcolor[HTML]{B7E1CD}1.10  & \cellcolor[HTML]{B7E1CD}0.34  & \multicolumn{2}{c|}{\cellcolor[HTML]{B7E1CD}3.22}  & \multicolumn{2}{c|}{\cellcolor[HTML]{B7E1CD}1.91}  & \multicolumn{2}{c|}{\cellcolor[HTML]{B7E1CD}0.49}  \\ \hline
\multicolumn{10}{|c|}{\textbf{SSL + DA}}                                                                                                                                                                                                                                                           \\ \hline
\textbf{B + 16W + Cutout}           & \cellcolor[HTML]{B7E1CD}1.54  & \cellcolor[HTML]{B7E1CD}1.27  & \cellcolor[HTML]{B7E1CD}0.43  & \multicolumn{2}{c|}{\cellcolor[HTML]{B7E1CD}2.17}  & \multicolumn{2}{c|}{\cellcolor[HTML]{B7E1CD}\textbf{2.81}}  & \multicolumn{2}{c|}{\cellcolor[HTML]{B7E1CD}1.02}  \\ \hline
\textbf{B + 16 W + box mixup}       & \cellcolor[HTML]{B7E1CD}1.2   & \cellcolor[HTML]{B7E1CD}1.67  & \cellcolor[HTML]{B7E1CD}1.66  & \multicolumn{2}{c|}{\cellcolor[HTML]{B7E1CD}1.42}  & \multicolumn{2}{c|}{\cellcolor[HTML]{B7E1CD}2.57}  & \multicolumn{2}{c|}{\cellcolor[HTML]{B7E1CD}\textbf{2.59}}  \\ \hline
\textbf{B + 16 W + boxmixup cutout} & \cellcolor[HTML]{B7E1CD} \textbf{3.51}  & \cellcolor[HTML]{B7E1CD} \textbf{1.84}  & \cellcolor[HTML]{B7E1CD}1.01  & \multicolumn{2}{c|}{\cellcolor[HTML]{B7E1CD} \textbf{5.57}}  & \multicolumn{2}{c|}{\cellcolor[HTML]{B7E1CD}2.53}  & \multicolumn{2}{c|}{\cellcolor[HTML]{B7E1CD}1.02}  \\ \hline
\textbf{B +16 W + cutpaste cutout}  & \cellcolor[HTML]{B7E1CD}2.87  & \cellcolor[HTML]{B7E1CD}1.38  & \cellcolor[HTML]{B7E1CD} \textbf{2.26}  & \multicolumn{2}{c|}{\cellcolor[HTML]{B7E1CD}5}     & \multicolumn{2}{c|}{\cellcolor[HTML]{B7E1CD}1.13}  & \multicolumn{2}{c|}{\cellcolor[HTML]{B7E1CD}1.19}  \\ \hline
\textbf{B +16 W + cutpaste}         & \cellcolor[HTML]{B7E1CD}0.98  & \cellcolor[HTML]{B7E1CD}0.67  & \cellcolor[HTML]{B7E1CD}0.72  & \multicolumn{2}{c|}{\cellcolor[HTML]{B7E1CD}1.61}  & \multicolumn{2}{c|}{\cellcolor[HTML]{B7E1CD}0.65}  & \multicolumn{2}{c|}{\cellcolor[HTML]{B7E1CD}0.73}  \\ \hline
\end{tabular}
\end{minipage}}
\end{table*}

We evaluate our models on the KITTI 3D detection benchmark which consists of 7,481 labeled training samples and 7518 unlabeled testing samples. Since the ground truth labels for the test set are not available, we evaluated our model by splitting the training set into 3711 training samples and 3768 validation samples. We experiment with ResNet-18 as the backbone. We implemented our deep neural network in Pytorch and trained using Adam optimizer with learning rate of 1.25*1e-4 for 200 epochs. We trained our network with a batch size of 16. Our model achieved best speed with 33 FPS on a NVIDIA GTX 2080Ti GPU. 

\textbf{Metrics :} The KITTI benchmark evaluates the models by Average Precision (AP) of each class (Car, Pedestrian and Cyclist).
We use the Mean Average Precision (mAP), the mean value of the Average Precision (AP) over all classes using equation (\ref{eq:3}) : 
\begin{equation}
    \text{mAP}_{3D} = \frac{1}{\left|C\right|} \sum_{c \in C} \text{AP}_c \label{eq:3}
\end{equation}
where $C=\{\text{car, pedestrian, cyclist}\}$.

\textbf{Inverse Class Frequency Weighted (ICFW) mAP} : We introduce a new metric that is used to demonstrate gains in a class imbalanced KITTI dataset. As the proposed metric is weighted by inverse of the class frequency, the gains over minority classed are favoured. The relative frequency (denoted by $f_c$ and in blue) of car, pedestrian and cyclist classes in the validation classes are shown in Table \ref{tab:classfreq}. This is evaluated by the following formula in equation (\ref{eq:4}): 
\begin{equation}
    w_c := \frac{f_c^{-1}}{\sum_{c \in C} f_c^{-1}} \in [0, 1] \text{ \ and \ } 
    \sum_{c \in C} w_c = 1   \label{eq:4}
\end{equation}
The values of $w_c$ are shown in Table \ref{tab:classfreq}.
Now the ICFW mAP is evaluated with equation (\ref{eq:5}) as:
\begin{equation}
    \text{ICFW mAP$_{3D}$} =  \sum_{c \in C} w_c \text{AP}_c \label{eq:5}
\end{equation}
Table \ref{tab:mAP} shows the results of DA and SSL evaluated on KITTI dataset. It tabulates the mAP2D, mAP3D \& mAPBEV over all classes for IoU=0.5. 

\textbf{SSL vs SSL+DA} : 
We observe an improved performance in both mAP$_{3D}$ and mAP$_{BEV}$ scores when using the MOL-SSL pretext task. In our study we evaluate the effect of 3 data augmentation schemes on the pretext task and thus the performance of the main task. They are detailed and demonstrated in Figure \ref{fig:data_aug_3}.
The combination of SSL-MOL along with cutout and box-mixup augmentations provide the largest of gains. This is attributed to the gains that cutout provide for truncated objects, and box-mixup for varied foreground/background variations.

\textbf{DA vs SSL+DA} : The proposed data augmentations alone without pretext tasks do also provide gains over the baseline while remaining lower in performance than combination SSL+DA. We hypothesize that DA provide augmented samples that help to improve the pretext task by imposing that the n/w should learn features to classify random windows in augmented samples as well as the primary 3D detection task. Cutout alone performed a bit worse over baseline, while in combination with the MOL pretext task, performance is greater than Cutout/SSL tasks alone.

\section{Conclusion}
In this study we evaluated the performance of the multi-object labeling pre-text task training in conjunction with the main monocular 3d-object detection task. As expected due to correlation between the classification task and the 3d localization task, we see an improvement of 1-2 points in the mAP scores for 3D and BEV metrics, besides the evident gains in mAP 2D scores. We also evaluated the performance of using data augmentation schemes along with the pre-text task, in our case we evaluated the performance of box-mixup (a bounding box version of instance mixup), cutout and box-cut-paste, and their combinations. We observed that the addition of data augmentation strategies improved the diversity of samples received by the MOL pretext task, and thus directly contributed in improving the performance of the main 3d detection task.


\appendix 
\section{Acknowledgements}
Authors would like to thank Navya to have provided the opportunity to perform applied research in key engineering problems.
\section{Class frequencies in KITTI-3D}
\begin{table}[h!]
\caption{Class freq. \& inverse weights on the validation set. }
\label{tab:classfreq}
\centering
\begin{tabular}{|l|c|c|c|}
\hline
\textbf{Class} & \textbf{Car} & \textbf{Pedestrian} & \textbf{Cyclist} \\
\hline
\textbf{Frequency $f_c$}     & 0.82 & 0.12 & 0.05 \\
\textbf{Inverted $w_c$}     & 0.04 & 0.27 & 0.69 \\
\hline
\end{tabular}
\end{table}

\section{Example results}
\begin{figure*}[hbt!]
\setlength{\lineskip}{0pt}
\centering
\begin{subfigure}[b]{0.45\textwidth}
\centering
    Baseline model\\
    \vspace{1mm}
    \includegraphics[width=\textwidth]{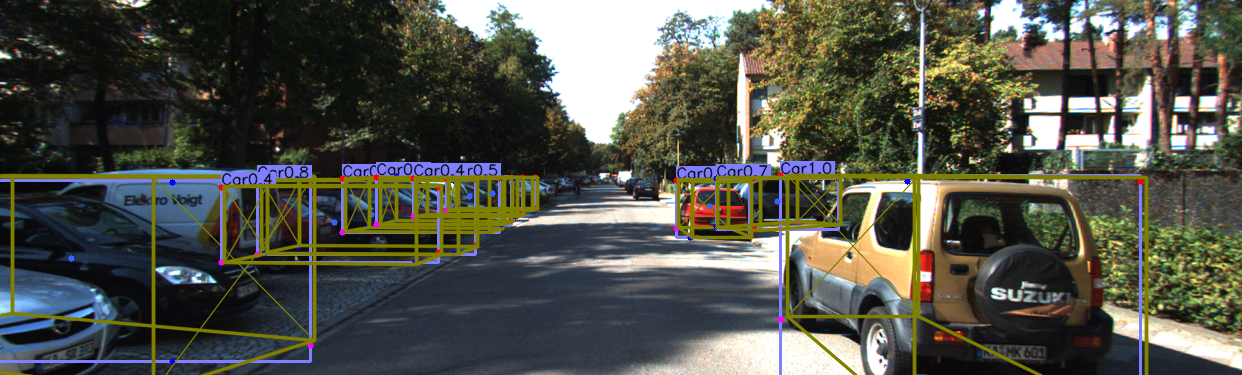}
    BoxMixup Augmentation\\
    \vspace{1mm}
    \includegraphics[width=\textwidth]{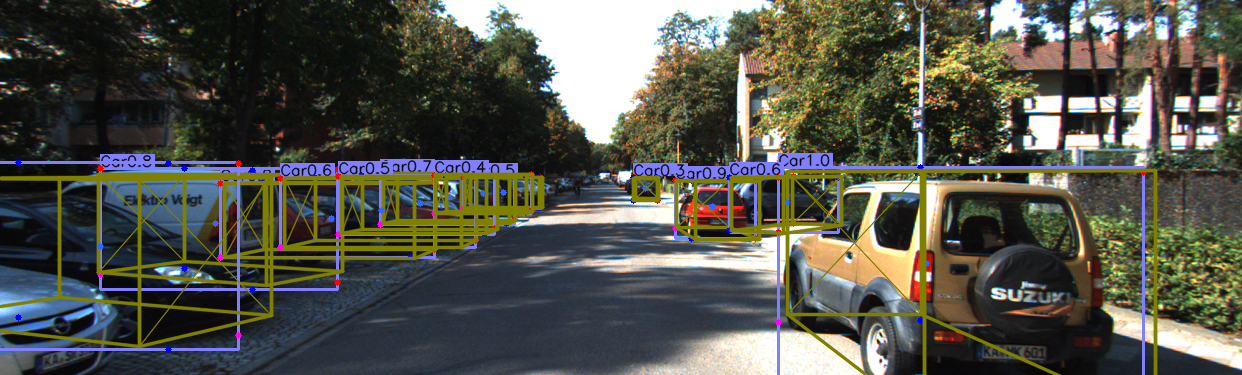}
\end{subfigure}
\hfill
\begin{subfigure}[b]{0.45\textwidth}
\centering
    \includegraphics[width=0.375\textwidth]{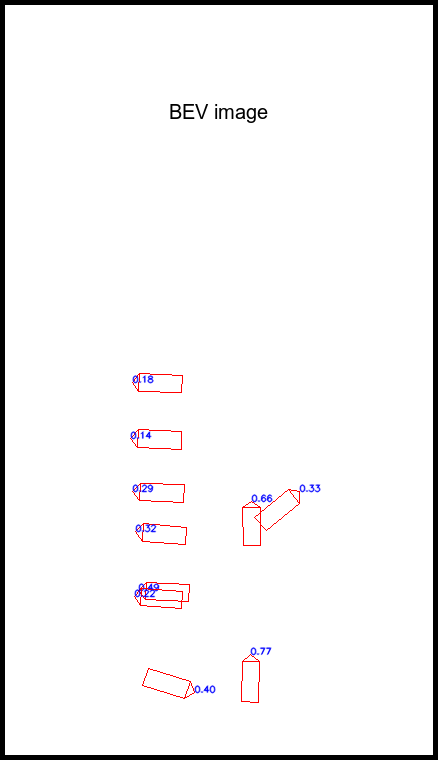}
    \includegraphics[width=0.375\textwidth]{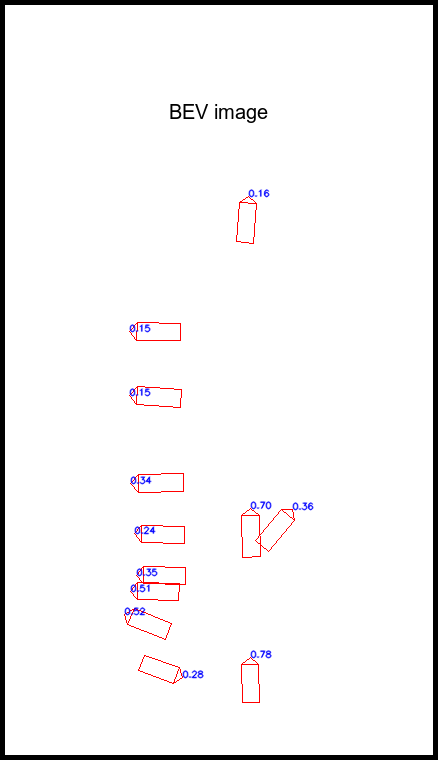}
    \caption{Left: Baseline, Right : Box-MixUp}
\end{subfigure}
\hfill
\begin{subfigure}[b]{0.45\textwidth}
\centering
    \vspace{1mm}
    Baseline model\\
    \vspace{1mm}
    \includegraphics[width=\textwidth]{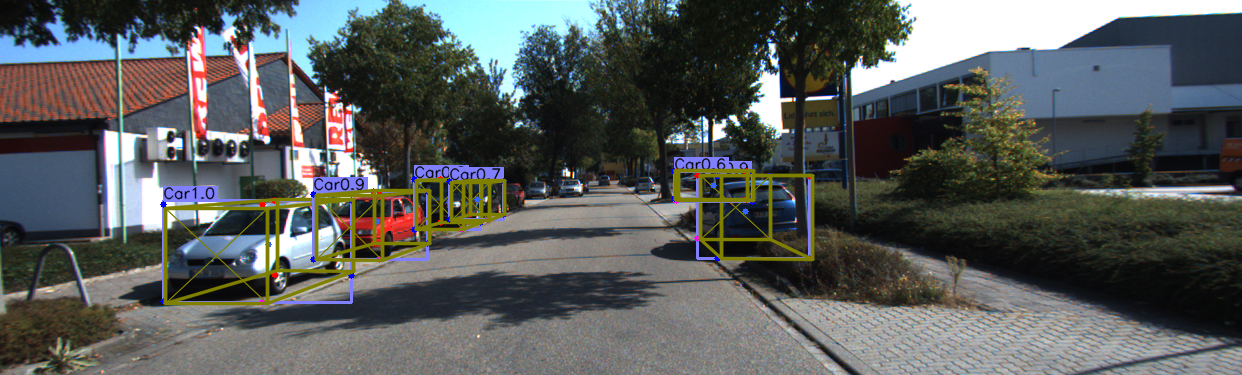}
    BoxMixup Augmentation\\
    \vspace{1mm}
    \includegraphics[width=\textwidth]{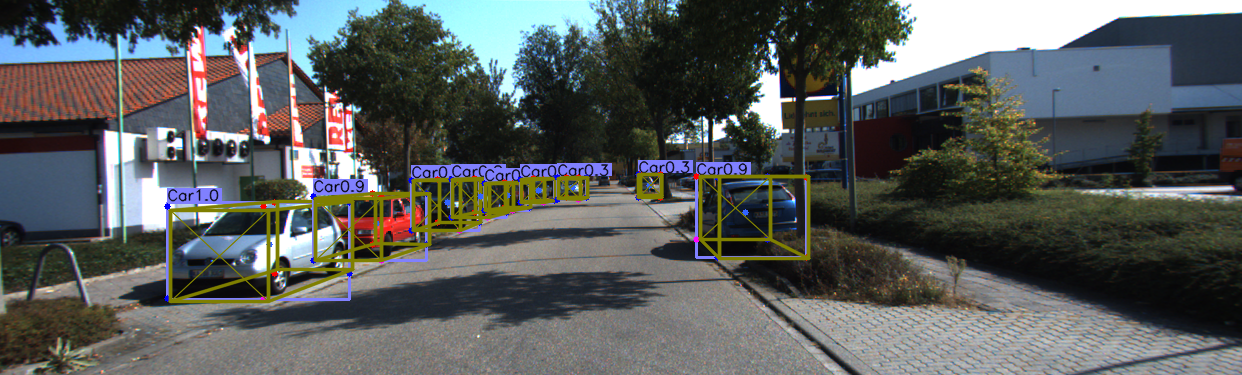}
\end{subfigure}
\hfill
\begin{subfigure}[b]{0.45\textwidth}
\centering
    \includegraphics[width=0.375\textwidth]{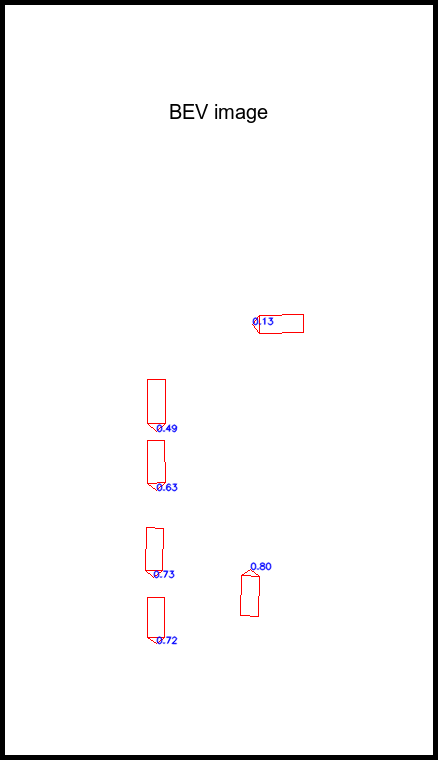}
    \includegraphics[width=0.375\textwidth]{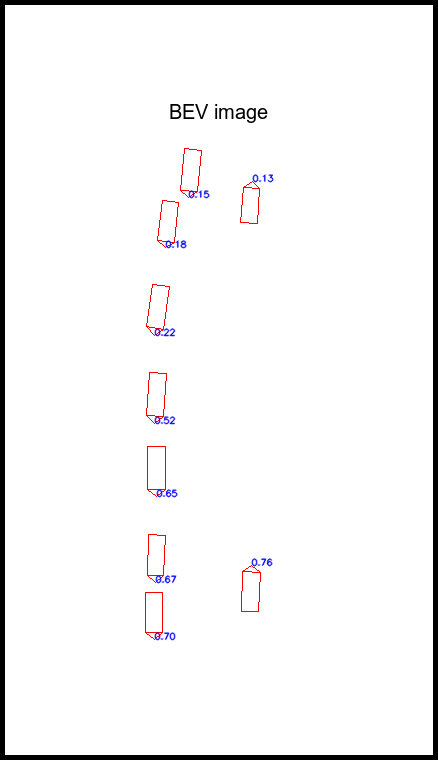}
    \caption{Left: Baseline, Right : Box-MixUp}
\end{subfigure}
\hfill
\begin{subfigure}[b]{0.45\textwidth}
\centering
    \vspace{1mm}    
    Baseline model\\
    \vspace{1mm}
    \includegraphics[width=\textwidth]{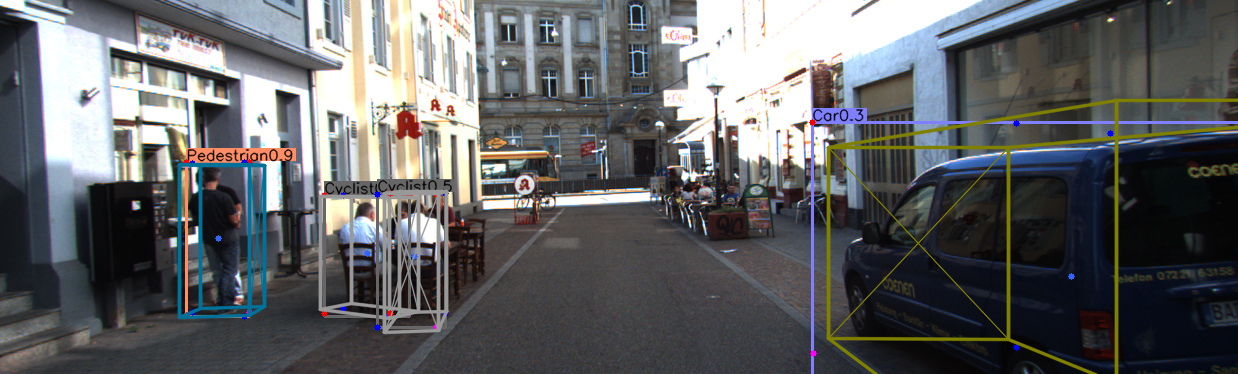}
    BoxMixup Augmentation\\
    \vspace{1mm}
    \includegraphics[width=\textwidth]{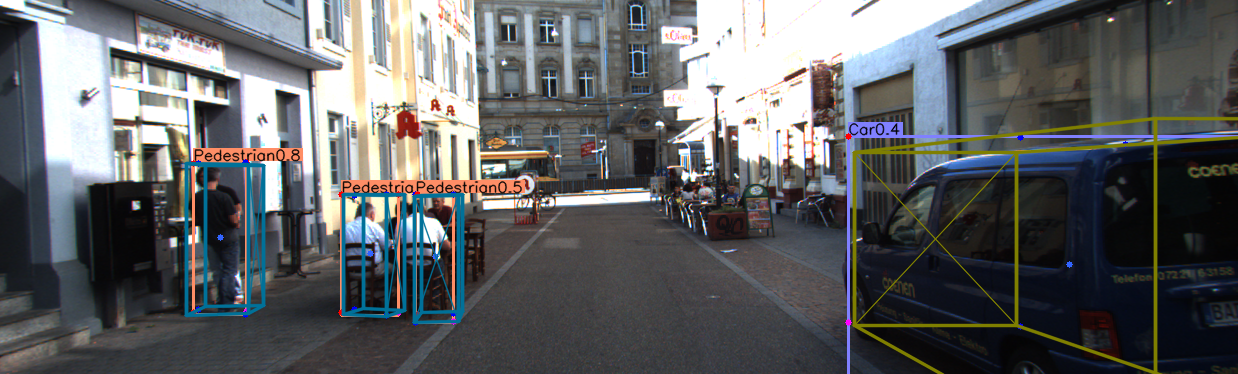}
\end{subfigure}
\hfill
\begin{subfigure}[b]{0.45\textwidth}
\centering
    \includegraphics[width=0.375\textwidth]{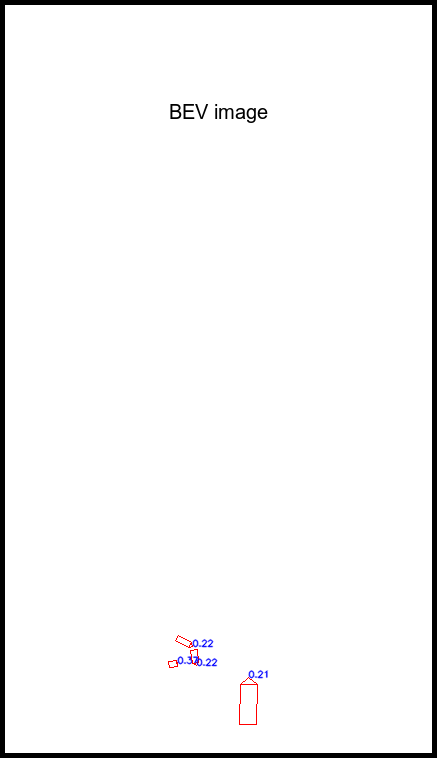}
    \includegraphics[width=0.375\textwidth]{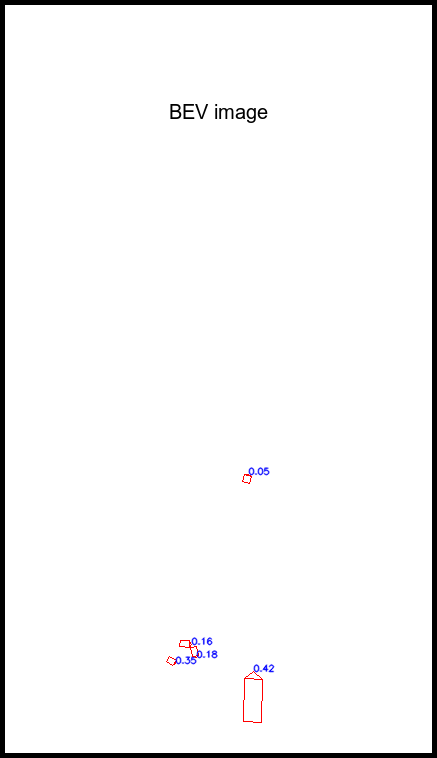}
    \caption{Left: Baseline, Right : Box-MixUp}
\end{subfigure}

\caption{Illustration of Box-Mixup data augmentation in various scenarios. Each time contains
the (baseline, Box-Mixup) prediction pair on the left panel, while the BEV representations (baseline, data augmented) pair on the right panel.}
\label{fig:img-pairs}
\end{figure*}

Fig. \ref{fig:img-pairs} shows the detection results of our proposed Box-Mixup data augmentation on KITTI in different scenarios. Ex. Occluded objects, missed detections and mis-classification by baseline. It shows baseline, Box-Mixup predictions on left panel and their corresponding BEV representations on right panel.

{\small
\bibliographystyle{ieee_fullname}
\bibliography{root}
}

\end{document}